\documentclass[a4paper,twoside]{article}

\usepackage{epsfig}
\usepackage{subcaption}
\usepackage{calc}
\usepackage{amssymb}
\usepackage{amstext}
\usepackage{amsmath}
\usepackage{amsthm}
\usepackage{multicol}
\usepackage{pslatex}
\usepackage{apalike}
\usepackage[bottom]{footmisc}
\usepackage[T1]{fontenc}
\usepackage[utf8]{inputenc}
\usepackage{authblk}

\usepackage{SCITEPRESS}     % Please add other packages that you may need BEFORE the SCITEPRESS.sty package.

\begin{document}

\title{DaDe: Delay-adaptive Detector for Streaming Perception}

\author[1]{Wonwoo Jo}
\author[2]{Kyungshin Lee}
\author[1]{Jaewon Baik}
\author[1]{Sangsun Lee}
\author[1]{Dongho Choi}
\author[1]{Hyunkyoo Park}
\affil[1]{\textit{C\&BIS Co.,Ltd.} \protect\\\textit{\small{$^*$\{wwjo, jwbaik, ssnlee, dhchoi, hkpark\}@cnbis.co.kr}}}
\affil[2]{\textit{\small{wawaworld23@gmail.com}}}


\abstract{Recognizing the surrounding environment at low latency is critical in autonomous driving. In real-time environment, surrounding environment changes when processing is over. Current detection models are incapable of dealing with changes in the environment that occur after processing. Streaming perception is proposed to assess the latency and accuracy of real-time video perception. However, additional problems arise in real-world applications due to limited hardware resources, high temperatures, and other factors. In this study, we develop a model that can reflect processing delays in real time and produce the most reasonable results. By incorporating the proposed feature queue and feature select module, the system gains the ability to forecast specific time steps without any additional computational costs. Our method is tested on the Argoverse-HD dataset. It achieves higher performance than the current state-of-the-art methods(2022.12) in various environments when delayed . The code is available at https://github.com/danjos95/DADE}

\onecolumn \maketitle \normalsize \vfill

\section{\uppercase{Introduction}}
\label{sec:introduction}
\begin{figure}[t]
\begin{center}
   \includegraphics[width=0.9\linewidth]{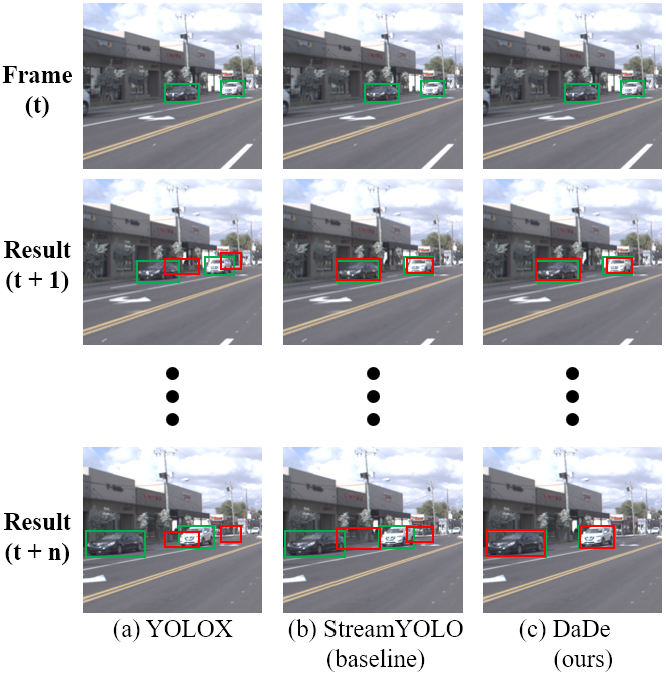}
\end{center}
   \caption{Visualization of the results of detectors (red boxes). Ground Truth (green boxes) changes due to image flow. Our method can detect objects correctly even after multiple time-step passes during processing.}
\label{fig:boxes}
\end{figure}

Recognizing the surrounding environment and reacting with low latency is critical in autonomous driving for a safe and comfortable driving experience. In practice, the gap between the input data and the surrounding environment widens as the processing latency increases. As shown in Figure \ref{fig:boxes}, surrounding environment changes from sensor input time $t$ to $t + n$ when the model finishes processing. To address this issue, some detectors \cite{yolov3}\cite{yolox} are focused on lowering the latency. They can complete the entire processing before the next sensor input. It appears reasonable but the processing delay caused a gap between the results and the changing environment. Furthermore, current image detection metrics such as average precision, and mean average precision do not consider a real-time online perception environment. It causes detectors to prioritize accuracy over the delay. To evaluate streaming performance, \cite{streamingperception} proposed a new metric, "streaming accuracy" to integrate latency and accuracy into a single metric for real-time online perception.
Strong detectors \cite{maskrcnn}\cite{cascadercnn} showed a significant performance drop in streaming perception. StreamYOLO \cite{streamyolo} created the model to predict future frames by combining the previous and current frames. It achieved state-of-the-art streaming AP performance by successfully detecting future frame-keeping real-time performance. 
In the real-world environment, processing time varies for various reasons(e.g. Temperature, Multi-modal Multi-task and sequential processing) while running the model. It requires detectors with the ability to perform multi-time-step predictions. For example, in Figure \ref{fig:boxes}(b), StreamYOLO outperforms the baseline detector in predicting the next time-step frame when the processing time is shorter than the inter-frame time. When processing time exceeds inter-frame time, the system begins to collapse and produces poor prediction results, as shown in Result $(t + n)$ in Figure \ref{fig:boxes}. The proposed system can function properly in an unexpectedly long processing time environment and provide the best detection at the time. 
In this study, we construct a model that provides insight into the required time-step frame. The feature queue module saves multi-time-step image features. It prevents the need for additional feature extraction processes, which incur additional computational costs. The dynamic feature select module constantly monitors the processing delay and chooses the best feature for the estimated delay. We use StreamYOLO's dual flow perception (DFP) module, which fuses the previous and current frames to make objects moving trend. The model can make moving trend of the target time step by fusing features from the dynamic feature select module.
Experiments are conducted on Argoverse-HD \cite{argoverse1.1} dataset. We tested delays in various settings. Our method showed significant improvement when compared to the baseline (StreamYOLO). As the mean processing delay increased, the difference between the baseline and ours widened.

This research makes three main contributions as follows:
\begin{itemize}
    \item We introduce delay-adaptive detector (DaDe), which can produce future results that are tailored to the output environment. Processing delay cannot be stable in a real-time environment and must be considered. We improved the baseline so that it could handle unexpected delays. This process has no computational cost or accuracy trade-off.
    \item We create a simple feature control module that can choose the best image feature based on the current delay trend. The feature queue module stores previous features to avoid additional computation. The feature select module monitors pipeline delays and chooses the best feature to create the moving trend. We can achieve accurate target time results by using the DFP module from StreamYOLO.
    \item Our system outperformed the state-of-the-art (StreamYOLO) model in delay-varying environments. We changed the mean delay time to stimulate delayed environments during the evaluation. Our method achieved $0.7\sim1.5\%$ higher sAP compared to the state-of-the-art method. This work also shows considerable room for a delay-critical system.
\end{itemize}

\begin{figure*}
\begin{center}
   \includegraphics[width=1\linewidth]{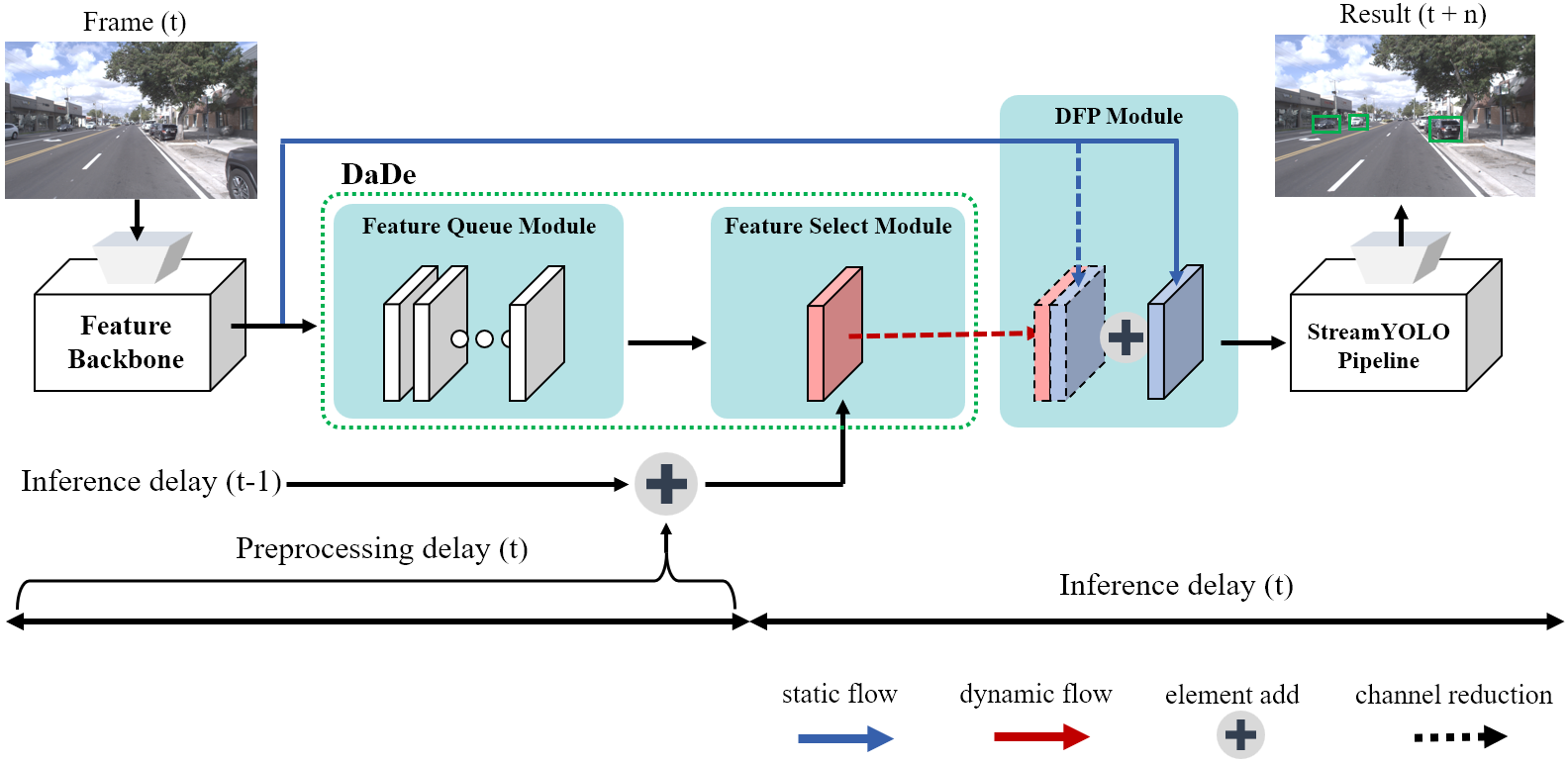}
\end{center}
   \caption{Delay-adaptive Detector is illustrated for delay adaptive stream perception. The image feature is extracted by CSPDarknet-53 and PANet, and stored in the fixed-length Feature Queue Module. The Feature Select Module investigates delay using the currently performed preprocessing delay and the most recent inference delay. The Feature Select Module chooses appropriate features from a list of stored features. Selected features are used to perform detection in the StreamYOLO pipeline.}
\label{fig:dade}
\end{figure*}

\section{\uppercase{Related Work}}
\subsection{Image object detection}
Both latency and accuracy are critical factors in object detection. Two-stage detectors \cite{rcnn}\cite{fastrcnn}\cite{maskrcnn}\cite{twostage_fpn} use a regional proposal system to focus on the accuracy of offline applications. However, in the real world, latency is more important than accuracy because the environment changes throughout processing. One-stage detectors \cite{onestage_focalloss_2017}\cite{onestage_iqdet_2021} have an advantage in processing time compared to regional proposed methods. YOLO series \cite{yolov3}\cite{yolox} is one of the mainstream performances in the one-stage detection method. Our research is based on the YOLOX \cite{yolox} detector. YOLOX added many advanced detection technologies (anchor free, decoupled head, etc.) from YOLOv3 \cite{yolov3} to achieve powerful performance.

\subsection{Video detection and prediction}
There have been attempts to improve the detection performance by using previous images from the video stream. Recent methods, such as \cite{videoobjectdetection_1}\cite{videoobjectdetection_2}\cite{videoobjectdetection_3}\cite{videoobjectdetection_4} use attention, optical flow, and a tracking method to aggregate image features and achieve high detection accuracy. The video future frame predictor \cite{videopredictionorig_1}\cite{videopredictionorig_2} creates unobserved future images from previous images. The ConvLSTM-based autoencoder \cite{videoprediction_1}\cite{videoprediction_2}\cite{videoprediction_3}\cite{videoprediction_4}\cite{videoprediction_5} generates representations of previous frames, and then the decoder generates future frame pixels based on those representations. However, they cannot be used in real-time applications because they are designed for an offline environment and do not account for latency.

\subsection{Progressive anytime algorithm}
There are previous studies on planning systems under resource constraints \cite{anytime_1_resourceconstraint} and flexible computation. The anytime algorithm can return results at any point in time. The quality of the results gradually increases as the processing time increases \cite{anytime_2}. However, the preceding studies do not consider environmental changes that occur during processing. It bridges the gap between the actual environment and the results. Our efforts are aimed at making the model more robust, even in delayed situations. It can produce appropriate results to the current time even when processing time becomes unexpectedly long.

\subsection{Streaming perception}
There are two types of detectors in a real-time system: real-time and non-real-time. A real-time detector can complete all processing steps before the next frame arrives. There was no metric to evaluate considering both latency and accuracy. As latency becomes more important in a real-world application, \cite{streamingperception} integrates latency and accuracy into a single metric. Streaming AP (sAP) was proposed to evaluate accuracy while considering time delay. It also suggests methods for recovering sAP performance drops, such as decision-theoretic scheduling, asynchronous tracking, and future forecasting. StreamYOLO \cite{streamyolo} reduced streaming perception to the task of creating a real-time detector that predicts the next frame. It achieved state-of-the-art performance, but when real-time processing is violated, the system can collapse because it can only produce the next time-step result.

\subsection{Delay variance}
If an additional delay occurs while processing, the gap between the results and the surrounding environment grows larger. This leads to a bad decision, which can lead to serious safety issues. Delays can occur for various reasons while running real-world applications.

\subsubsection{Temperature} Temperature is a well-known issue in a real-time system. Performance may be reduced to avoid permanent damage to the processing chips. In our hardware environment, processing speed drops to $70\%$ at $90 ^\circ C$, with a performance drop of up to $50\%$ at $100 ^\circ C$. In autonomous driving, the temperature control gets harder since the device operates in various outdoor environments.

\subsubsection{Multiple sensor configuration} It is critical to use multiple sensors to perceive the surrounding environment to obtain diverse data or a multi-view of the environment. It is becoming more common to use lidar-camera fusion \cite{multisensorbevfusion}\cite{fusion_lidarcamera_mv3d} or multi-camera system \cite{multicamerabevformer}\cite{multicamera_detr3d} to make surrounding recognition. Each sensor has a unique frequency and time delay. Here, fast sensor data must wait until all sensor data arrives, which can add to the delay.

\subsubsection{Multiple model deploy} To build an autonomous driving system, multiple tasks (for example, lane segmentation, traffic light detection, 3D object detection, multi-object tracking, and so on) must be performed parallel and dynamically. \cite{Multiplemodel_1} proposes a multiple detection model with a shared backbone to reduce computational costs. In any case, running multiple models can cause an unexpected delay in the system's sequential task.
If the processing delay exceeds the inter-frame spacing, the real-time system fails and the detector results become out-of-date. We changed the mean delay time to stimulate delayed environments during the evaluation.

\subsection{Preprocessing and inference delay}
There are preprocessing and inference stages in the processing sensor system. The main processing unit receives raw sensor data and refines it to create a proper shape for the model during the preprocessing stage. Following preprocessing, input data flows into a neural processing unit, which produces results through neural networks. Since it runs sequentially, preprocessing is fast but the delay can be easily changed by a processing bottleneck. As the task becomes more demanding, the number of sequential operations in the main processing unit increases, which can directly affect the delay.
\begin{figure}[tb!]
\begin{center}
   \includegraphics[width=0.88\linewidth]{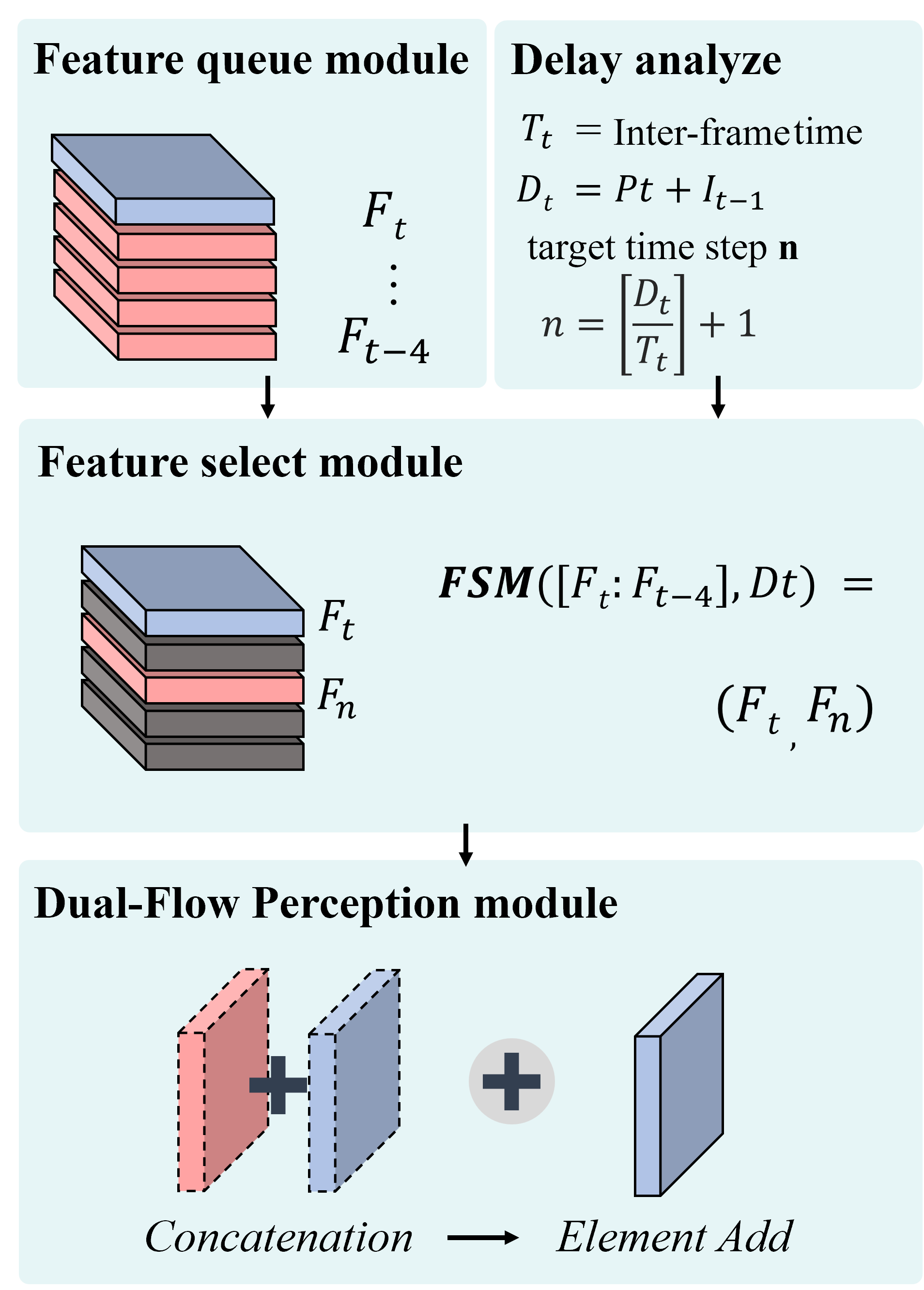}
\end{center}
   \caption{Pipeline of DaDe. The feature queue of the module stores previous image features to avoid additional computation. The feature select module selects features from queues based on the current delay time. The dual flow perception module generates object moving trends based on input features.}
\label{fig:module}
\end{figure}
\section{\uppercase{Methodology}}
In this study, we present a simple and effective method for achieving the desired future results. As in Figure \ref{fig:dade}, the Delay-adaptive Detector (DaDe) can return the desired time-step by equipping the feature queue module, feature select module, and Dual-Flow Perception (DFP) module. We avoid adding extra computation in processing to maintain real-time performance.

\subsection{Dual Flow Perception module}
StreamYOLO \cite{streamyolo} suggested a DFP module for tracking moving trends. DFP uses the previous frame's feature map. Dynamic flow in DFP fuses the features of two adjacent frames to create an object moving trend, whereas static flow is created using a standard feature extraction based on the current frame only. 1x1 convolution layer is used for dynamic flow, which is followed by the batchnorm and SiLU \cite{batchnormandSiLU} to reduce the channel for previous and current features in half. Then, DFP concatenates these two reduced features to generate the dynamic features. The detector can sense the object's moving trends and detect future frames without additional latency by combining dynamic flow and static flow. To advance this method, we use a multi-time-step feature fusing system to create a delay adaptive detector. The model can make a multi-time-step prediction by following the feature queue and feature select module.

\subsection{Feature queue module}
The feature queue module stores extracted image features in a featured queue of a fixed length. Features are represented as $F_t$, whereas $t$ is the corresponding time-step. The computational cost of saving old image features is as same as StreamYOLO. The feature queue module continues to save features from the four previous frames. Storing the previous feature prevents additional processing while occupying little additional memory (100MB for each frame feature). This enables the model to use older time-step features.

\subsection{Delay analyze}
Preprocessing and inference time in time step $t$ are denoted as $P_t$ and $I_t$ respectively. The inter-frame time between input frames is denoted by the symbol $T_t$. The delay trend can be calculated by adding the most recent inference delay and the current preprocessing delay. The final delay trend $D_t$ is as follows:
\begin{equation}
D_t  =P_t+I_{(t-1)}
\label{eqn:1}
\end{equation}
\subsection{Feature select module}
The feature select module chooses two time-step features that best match the current delay trend.
When the delay trend is entered into the feature select module, the target time-step $n$ can be calculated using Equation \ref{eqn:2}. The expected time-step is the delay trend quotient over inter-frame time. The target time-step $n$ is the next time-step of the division result.
\begin{equation}
n =  \left[\frac{D_t}{T_t}\right]+1
\label{eqn:2}
\end{equation}

In the feature select module, features for the required time step are extracted as the equation below.
\begin{equation}
\textbf{FSM}([F_t:F_{(t-4)}], D_t) = (F_t , F_{(t-n)})
\label{eqn:3}
\end{equation}
If there is no last inference delay, the result of FSM should be $(F_t , F_{t-1})$. If no suitable feature exists for $n$, the nearest $F$ should be chosen.
In summary, the proposed model calculates the time delay by combining the current preprocessing delay and the last inference delay. After determining the desired time-step using Equation \ref{eqn:2}, the module chooses the stored features using Equation \ref{eqn:3}. The DFP creates a dynamic time-step object moving trend using the selected feature. The entire feature flow process is explained in Figure \ref{fig:module}.

\section{\uppercase{Experiments}}

\subsection{Environment settings}
\subsubsection{Dataset}Argoverse-HD \cite{argoverse1.1} provides high-resolution (1920x1080) and high-frame-rate driving video image data (30FPS). The Argoverse-HD validation set contains 24 videos of $15\sim30$ seconds each, totaling 15k image frames. Streaming perception \cite{streamingperception} adds dense 2D annotations in MS COCO \cite{coco} format.

\subsubsection{Streaming AP}Streaming AP \cite{streamingperception} is used in the evaluation of experiments. During evaluation, the model updates the output buffer with its latest prediction of the current state of the world. At the same time, the benchmark constantly queries the output buffer for estimates of world states. So, output will be evaluated with the current ground truth frame in world state. To  AP metric from COCO \cite{coco}, sAP evaluates the average mAP over intersection-over-union (IoU) thresholds at 0.5 and 0.75. sAP$_S$, sAP$_M$, and sAP$_L$ denote sAP for object size. 

\subsubsection{Platforms} Each experiment in this study was run on an RTX 3070Ti GPU (8GB VRAM, TDP 80W), an Intel 12700 H CPU, and 16 GB RAM. Pytorch 1.7.1 with CUDA 11.4 was used in the software environment.

\begin{table}[tb!]
{\small{
%height blank adjustment
\renewcommand{\arraystretch}{1.3}
\resizebox{\columnwidth}{!}{%
\begin{tabular}{ccccc}
\hline
Environment &
  \begin{tabular}[c]{@{}c@{}}Mean\\ delay\end{tabular} &
  \begin{tabular}[c]{@{}c@{}}Standard\\ deviation\end{tabular} &
  \begin{tabular}[c]{@{}c@{}}Minimum\\ delay\end{tabular} &
  \begin{tabular}[c]{@{}c@{}}Maximum\\ delay\end{tabular} \\ \hline
\multicolumn{5}{c}{\textbf{Low}}    \\ \hline
StreamYOLO  & 23.5  & 3.2   & 21.8 & 69.1 \\
DADE(Ours)  & 24.1  & 3.66  & 21.9 & 66   \\ \hline
\multicolumn{5}{c}{\textbf{Medium}} \\ \hline
StreamYOLO  & 40.1  & 9.35  & 22.3 & 86.8 \\
DADE(Ours)  & 39.3  & 9.22  & 22.3 & 88   \\ \hline
\multicolumn{5}{c}{\textbf{High}}  \\ \hline
StreamYOLO  & 63    & 12.5  & 41.7 & 121  \\
DADE(Ours)  & 63.1  & 12.7  & 41.3 & 124  \\ \hline
\end{tabular}%
}}}
\caption{Delay analysis in various delay environments. All delays are measured in milliseconds.}
\label{table:delay}
\end{table}

\subsubsection{Delay} 
\begin{figure}[tb!]
\begin{center}
   \includegraphics[width=0.78\linewidth]{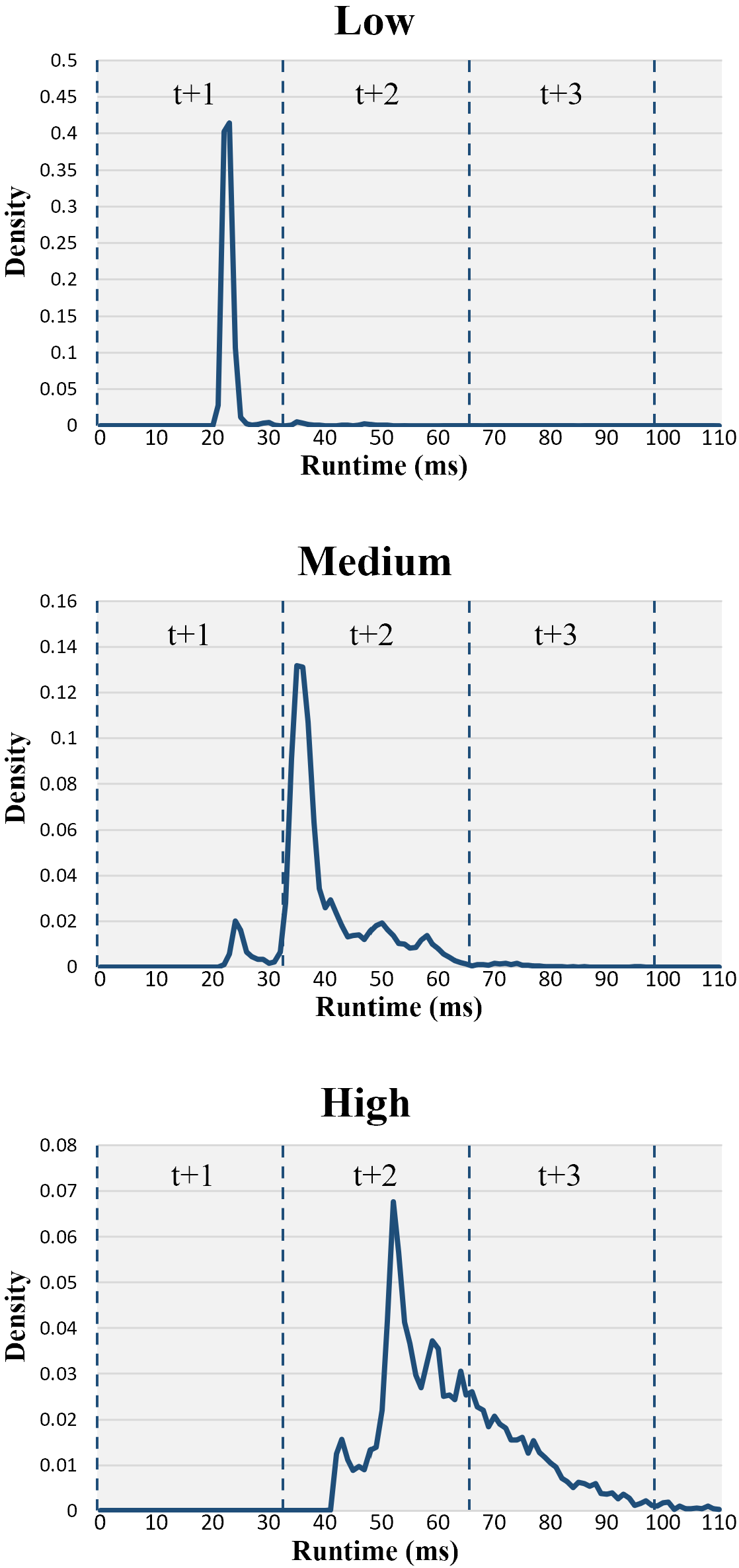}
\end{center}
   \caption{Visualization of delay distribution for various delay environments. Distribution gradually spreads as the mean delay lengthens.}
\label{fig:delaydistributionsvertical}
\end{figure}

To change mean delay time, we deployed multiple DaDe model on single environment. Table \ref{table:delay} shows various delay environments (Low, Medium, and High) by running up to 3 models in parallel. Inter-frame space is $33ms$ at 30FPS, so processing time should be less than $33ms$ to maintain real-time performance. In a low delay environment, the average delay is $24ms$, so StreamYOLO and DaDe can process each frame before the next frame arrives. In a medium delay environment, the average delay is $40ms$, after which some frames fail to meet real-time performance. As a result, their evaluation is based on frame$(t + 2)$ rather than frame$(t + 1)$. The mean delay time in a high delay environment is more than $60ms$, so the difference between the assumed future frame and the real environment increases in the baseline. Figure \ref{fig:delaydistributionsvertical} shows the processing delay distribution in each delay environment. A vertical dotted line shows inter-frame spaces at 30 frames per second (FPS). Frames in each inter-frame space should produce corresponding time-step results. 
\subsection{Results}
\begin{figure*}[t]
\begin{center}
   \includegraphics[width=1\linewidth]{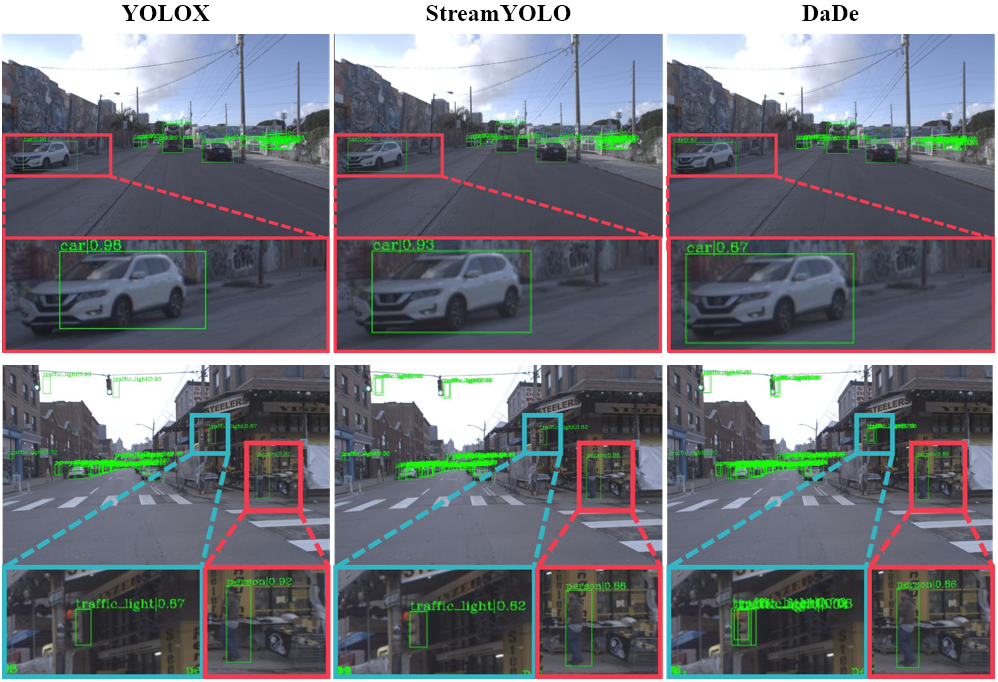}
\end{center}
   \caption{Qualitative analysis of results. Detection is conducted in a streaming environment with a high delay. DaDe can successfully predict the future environment by forming object-moving trends. It achieves better results not only for moving objects (top row) but also for detecting objects during ego-vehicle turns (bottom row).}
\label{fig:examples}
\end{figure*}

Table \ref{table:APresults} shows the results of streaming perception methods. The performance is the same as in a low-delay environment between baseline (StreamYOLO) and ours (DaDe). Standard YOLOX suffers from a significant performance drop in the absence of future prediction. The baseline fails to follow the trend of future frames as the mean delay increases and the frame begins to fail to finish processing before the next frame arrives. In high-delay conditions, all frames fail to detect in real-time, increasing the gap between the baseline and the suggested method. As shown in Table \ref{table:APresults}, a 1.5\% improvement in sAP is achieved compared to the baseline in a high delay environment. Figure \ref{fig:examples} shows the visualization of each method in a high delay environment. Ours outperforms others in various conditions and scenarios.
\begin{table}[tb!]
{\small{
%height blank adjustment
\renewcommand{\arraystretch}{1.4}
\resizebox{\columnwidth}{!}{%
\begin{tabular}{lcccccc}
\hline
Method              & sAP           & sAP$_L$         & sAP$_M$         & sAP$_S$         & sAP$_{50}$        & sAP$_{75}$        \\ \hline
\multicolumn{7}{c}{\textbf{Low} (mean delay: $24\pm1$ms)}                                                                     \\ \hline
YOLOX          & 31.7          & 52.3          & 29.7          & 12.5          & 54.8          & 30.1          \\
StreamYOLO          & \textbf{36.7}          & 63.8          & \textbf{37.0}          & \textbf{14.6}          & \textbf{57.9}          & 37.2          \\
\textbf{DADE(Ours)} & \textbf{36.7} & \textbf{63.9} & 36.9 & \textbf{14.6} & \textbf{57.9} & \textbf{37.3} \\ \hline
\multicolumn{7}{c}{\textbf{Medium} (mean delay: $39\pm1$ms)}                                                                  \\ \hline
YOLOX          & 23.0          & 40.0          & 21.3          & 6.1          & 45.5          & 21.1          \\
StreamYOLO          & 29.1          & 48.1          & 27.4          & 9.5          & 51.6          & 27.7          \\
\textbf{DADE(Ours)} & \textbf{29.8} & \textbf{50.1} & \textbf{28.3} & \textbf{10.8} & \textbf{52.5} & \textbf{28.4} \\ \hline
\multicolumn{7}{c}{\textbf{High} (mean delay: $63\pm1$ms)}                                                                    \\ \hline
YOLOX          & 19.2          & 31.2          & 15.5          & 5.1          & 38.8          & 16.3          \\
StreamYOLO          & 25.7          & 40.2          & 23.9          & 8.6           & 47.3          & 23.7          \\
\textbf{DADE(Ours)} & \textbf{27.2} & \textbf{41.5} & \textbf{26.1} & \textbf{9.5}  & \textbf{48.2} & \textbf{25.9} \\ \hline
\end{tabular}
}}}
\caption{Performance of each method in Argoverse-HD dataset. YOLOX is the standard real-time detector.}
\label{table:APresults}
\end{table}

\section{\uppercase{Conclusions}}
This study proposes a simple but effective method for detecting objects in real time. Multi-time-step prediction can be performed using the feature queue module and feature select module without any additional computation. It achieved state-of-the-art performance in a delayed environment. Furthermore, real-time feature operation in DaDe can significantly improve the performance of other real-world object detection tasks. Predicting the appropriate time step can significantly improve their localization accuracy. For further research, only one time-step feature is considered in our method to generate the object's moving trend. Accuracy can be improved if each time-step feature is considered within a reasonable computation cost. 

\vfill
%\section*{\uppercase{Acknowledgements}}
%This work was supported by 

\bibliographystyle{apalike}
{\small
\bibliography{example}}

\end{document}